\def\year{2021}\relax
\documentclass[letterpaper]{article} 
\usepackage{aaai21}  
\usepackage{times}  
\usepackage{helvet} 
\usepackage{courier}  
\usepackage[hyphens]{url}  
\usepackage{graphicx} 
\usepackage[switch]{lineno}
\usepackage{natbib}  
\usepackage{caption} 
\nocopyright

\title{Automated Source Code Generation and Auto-completion Using Deep Learning: Comparing and Discussing Current Language-Model-Related Approaches}
\author{
    Juan Cruz-Benito\textsuperscript{\rm 1}\thanks{Contact author}
    Sanjay Vishwakarma\textsuperscript{\rm 2}\thanks{Intern at IBM Quantum at the time of writing this paper}, 
    Francisco Martin-Fernandez\textsuperscript{\rm 1},
    Ismael Faro\textsuperscript{\rm 1},
    \\
}
\affiliations{

    \textsuperscript{\rm 1}IBM Quantum. IBM T.J. Watson Research Center. \\Yorktown Heights, NY 10598, USA\\
    
    \textsuperscript{\rm 2}Electrical and Computer Engineering. Carnegie Mellon University. \\Mountain View, USA\\
    juan.cruz@ibm.com, svishwak@andrew.cmu.edu, paco@ibm.com, ismael.faro1@ibm.com

}

\begin{document}

\maketitle

\begin{abstract}
In recent years, the use of deep learning in language models gained much attention. Some research projects claim that they can generate text that can be interpreted as human-writing, enabling new possibilities in many application areas. Among the different areas related to language processing, one of the most notable in applying this type of modeling is programming languages. For years, the Machine Learning community has been researching this software engineering area, pursuing goals like applying different approaches to auto-complete, generate, fix, or evaluate code programmed by humans. Considering the increasing popularity of the Deep-Learning-enabled language models approach, we detected a lack of empirical papers that compare different deep learning architectures to create and use language models based on programming code. This paper compares different neural network architectures like AWD-LSTMs, AWD-QRNNs, and Transformer while using transfer learning and different tokenizations to see how they behave in building language models using a Python dataset for code generation and filling mask tasks. Considering the results, we discuss each approach’s different strengths and weaknesses and what gaps we find to evaluate the language models or apply them in a real programming context. 
\end{abstract}

\section{Introduction}

\noindent We are digitally surrounded by computational Language Models (LMs) that guide us while writing to reduce the user effort, suggest different options for words/sentences to enhance our style, or fix our grammatical/correctness errors accurately \cite{kannan2016smart,bryant2018language, ghosh2017neural}. Many of the keys we press while writing on a keyboard act as part of the inputs to compose new datasets for those models that shape how we communicate with others. Nevertheless, does it happen in the same way when we write code? 
Succinctly, yes. According to some recent surveys found in the literature  \cite{allamanis2018survey,chen2020deep}, the Natural Language Processing (NLP) subfield related to programming language includes examples of LMs used in several tasks and contexts. For example, the authors of \cite{nguyen2015graph,bielik2016phog,cruz2018deep} used different techniques such as graph-based statistical LMs, probabilistic LMs, or Deep Learning (DL) LMs to suggest code to programmers similarly to auto-completer features in IDEs. LMs were used to generate automated source code based on sample code inputs or pseudo-code and evaluating how this generated code performs \cite{oda2015learning,tiwang2019deep,fedus2018maskgan}. Another exciting application of NLP into source code languages is the automatic translation between different languages. The work reported in \cite{nguyen2013lexical,roziere2020unsupervised} explores different supervised and unsupervised approaches to migrate code between different programming languages to improve interoperability or port codebases written in obsolete or deprecated languages (such as COBOL or Python2). Another example found is the use of Bayesian networks, attention mechanisms, and pointer networks \cite{proksch2015intelligent,li2017code, donahue-etal-2020-enabling} to fill a given code portion with missings.

There is a more general understanding of the natural languages’ different characteristics in the NLP broad field. Since there exist many research fields related to human languages, there is a richer background on existing language characteristics. For example, there is much knowledge on aspects like the minimal representation units of a word in a specific language, the most used words of a language, or if a word is a neologism or not. Programming languages share some syntax similarities with spoken languages. However, it does not have the same restrictions in the sense of common words or neologisms \cite{allamanis2015suggesting, karampatsis2019maybe}, or other syntax restrictions and features such as punctuation, format, or style. Every programming language has indeed reserved words and symbols to denote different actions, resources, or syntax. However, there is an essential part of the source code that is only limited by the programmer’s imagination, the conventions existing, or the guides for good practices. As \cite{karampatsis2019maybe} claims, 

\begin{quote}
[...] traditional language models limit the vocabulary to a fixed set of common words. For code, this strong assumption has been shown to have a significant negative effect on predictive performance [...]\end{quote}

In that paper, Karampatsis and Sutton \citeyear{karampatsis2019maybe} present how segmenting words into subword units can improve source code modeling. Similarly, other researchers \cite{ganin2016domain,kimcharacter2016,karpathy2016} dug in representing source code vocabulary with a similar emphasis on modeling words using sub-word units and envisioning their importance when using neural networks (NNs). Nevertheless, how that word segmentation affect the accuracy or the appropriateness of the code generated or auto-completed in some modern LM using deep learning approaches? That kind of question raises the main goal for this paper: discover what kinds of associations between different modern neural network architectures and tokenization models produce the best results when creating LMs to generate and auto-complete source code.

To pursue that goal, this research aims to conduct experiments combining different deep neural network (DNN) architectures with different tokenization and pre-trained models over an existing Python dataset. Using that experimentation, we want to investigate the combinations that improve code generation and auto-completion tasks (for example, filling the blanks) while checking the outcomes from those tasks using metrics like accuracy and human assessment.

The rest of the paper is as follows: Section 2 presents the different approaches followed during the research, the DNNs used, the software methods and data employed. Section 3 describes results achieved during the research according to different metrics and tests, while section 4 discusses these findings and the implications of the results as appropriate. Finally, Section 5 presents some conclusions.

\section{Materials and methods}

We have trained a set of deep neural networks using different architectures, tokenization techniques, and software libraries to develop this research work. Following, we introduce the different materials and methods employed for that purpose.

\subsection{Deep Neural Networks and tokenization models used}
Regarding the DNN architectures employed, we chose to use the following ones: ASGD Weight-Dropped LSTM (AWD-LSTM) \cite{merity2018regularizing}, Quasi Recurrent Neural Networks (QRNNs) \cite{bradbury2016quasi}, and Transformer \cite{vaswani2017attention}. These DNN architectures have been reportedly getting some state-of-the-art results \cite{merity2018regularizing,pmlr-v97-wang19f,gong2018frage,takase2018direct,yang2018breaking,pmlr-v80-krause18a,rae2019compressive,dai2019transformer,baevski2018adaptive,merity2018analysis,howard2018universal,eisenschlos-etal-2019-multifit,brown2020language, devlin2019bert} recently in the NLP field in several groundbreaking digital products\footnote{https://openai.com/blog/openai-api/}$^{,}$\footnote{https://blog.google/products/search/search-language-understanding-bert/} and in some of the most known datasets like the Penn Tree Bank \cite{mikolov2011empirical}, WikiText-2 and WikiText-103 \cite{merity2016pointer}, the One-Billion Word benchmark \cite{chelba2014one}, or The Hutter Prize Wikipedia dataset \footnote{http://prize.hutter1.net/}. 

The AWD-LSTM is a variation of the famous Long Short-Term Memory (LSTM) architecture \cite{hochreiter1997long}. The LSTM is a type of Recurrent Neural Network (RNN) especially capable of processing and predicting sequences. That ability with sequences is the reason why LSTMs have been employed largely in LMs \cite{kimcharacter2016}. The AWD-LSTM includes several optimizations compared to the regular LSTM. Two of the most important ones are the use of Average Stochastic Gradient Descent (ASGD) and the WeightDropout. The ASGD is used as the NN’s optimizer to consider the previous weights (not only the current one) during each training iteration. The WeightDropout introduces the dropout technique \cite{srivastava2014dropout} to avoid overfitting but with the characteristic of returning zero, not with a subset of activations between layers, like in traditional dropout, but with a random subset of weights. 

The QRNN is another type of RNN that includes alternate convolutional and pooling layers in the architecture. This design makes the QRNN able to capture better long-term sequences and train much faster since the convolutional layers allow the computation of intermediate representations from the input in parallel. They can be up to 16 times faster at train and test time than LSTMs while having better predictive accuracy than stacked LSTMs of the same hidden size. We use a QRNN modified (AWD-QRNN) to include the same ASGD and WeightDropout modifications to improve its stability and optimize its capabilities, as for the AWD-LSTM.

We utilize AWD-LSTM and AWD-QRNN to produce LMs capable of solving the task of generating source code based on input as in the literature \cite{merity2018analysis,merity2018regularizing,pmlr-v97-wang19f,gong2018frage,takase2018direct,yang2018breaking,pmlr-v80-krause18a,merity2018analysis,howard2018universal,eisenschlos-etal-2019-multifit}. 

The Transformer is probably the most popular current DNN architecture in NLP due to its performance and recent state-of-the-art results in many tasks. It is an encoder-decoder architecture in which each layer uses attention mechanisms. This use of (self-)attention mechanisms makes Transformer able to model the relationships between all words in a sentence regardless of their respective position. That implies a significant improvement over RNNs, enabling much more parallelization of data processing and unblocking the training over more massive datasets. The excellent results of the Transformer architecture empowered the NLP community to create new state-of-the-art transformer-based models \cite{young2018recent} like those used in the current research: GPT-2 \cite{radford2019language}, BERT \cite{devlin2019bert}, and RoBERTa \cite{liu2019roberta}.

We choose to use GPT-2 since it is a causal transformer (unidirectional) that can predict the next token in a sequence. So, it can generate source code based on input, allowing us to compare the results with the AWD-LSTM and AWD-QRNN experiments. Regarding BERT and RoBERTA, we used them to study how a masked modeling approach can perform auto-complete source code. In that case, we did not use them for text generation, as in the other experiments, since BERT and RoBERTa are not designed for text generation. However, they can generate text (more diverse but slightly worse in quality) \cite{wang2019bert}.

Considering the tokenization techniques, for every AWD-LSTM and AWD-QRNN, we chose the following types of tokens: word, unigram, char, and byte-pair encoding (BPE) \cite{sennrich2015neural} -albeit some studies show that BPE is suboptimal for pre-training \cite{bostrom2020byte}-. For the Transformer models, we used the default ones from the pre-defined models: Wordpiece method \cite{schuster2012japanese} for BERT and BPE over raw bytes instead of Unicode characters for GPT-2 and RoBERTa. The different techniques were selected because they produce different tokens’ granularity that can enrich our experimentation: full words, sub-words of specific sizes, character-sized tokens, or byte pairs. Also, they enable us to compare the tokenization between the different types of models and tasks to solve.

\subsection{Experimentation details}

The dataset used for the experimentation is the Python dataset included in the “GitHub CodeSearchNet Challenge dataset” \cite{husain2019codesearchnet}. It includes 2 million (comment, Python code) pairs from open-source libraries. As observed during the dataset preparation for our experiments, there are about 11 million Python code sentences. The reason to choose this dataset is that it has already been used in previous research related to NLP and source code. The full dataset includes several languages: Go, Java, JavaScript, PHP, Python, and Ruby. We chose to use only the Python part of the dataset because it enables us to compare the existing literature that uses Python language more than other programming languages. The software libraries and packages used primarily during the research were the following: FastAI \cite{howard2020fastai}, Google SentencePiece \cite{kudo2018sentencepiece}, and HuggingFace's Transformers \cite{Wolf2019HuggingFacesTS}. 
he preprocessing applied to the dataset included removing most of the code comments and autoformatting the code according to the PEP-8 Python style guide using the \textit{autopep8}\footnote{https://pypi.org/project/autopep8/} package. Regarding the AWD-LSTM networks, we have been using the FastAI-provided base models pre-trained using the Wikitext-103 dataset \cite{merity2016pointer}. There are no default pre-trained models in the FastAI’s AWD-QRNN version of those networks, so we trained them from scratch. As we introduced, regarding the Transformer architectures, we have been using three standard pre-trained models as a basis: GPT-2, BERT, and RoBERTa. In each case, the exact pre-trained model used were \textit{gpt2}, \textit{bert-base-cased}, and \textit{roberta-base}. These pre-trained models are available from HuggingFace’s model\footnote{https://huggingface.co/models}. 

As the reader could infer from the previous explanations about using pre-trained versions, we followed a transfer learning approach similar to other researchers in existing literature \cite{howard2018universal,ruder2019transfer,chronopoulou2019embarrassingly,eisenschlos-etal-2019-multifit}. We employed the pre-trained models on English texts to later fine-tune the models for the selected tasks using the GitHub CodeSearchNet dataset.
The deep-neural-networks-related source code was coded using the FastAI library (versions 1.0.61 and 2 \textit{dev 0.0.21}). To apply the different tokenization techniques into the AWD-LSTMs and QRNNs, we replaced the default Spacy tokenizer \cite{honnibal2017spacy} with Google Sentencepiece \cite{kudo2018sentencepiece}, following a similar approach to \cite{czapla2018universal}. In the development of Transformer architectures to see how they perform filling the blanks and generating texts, we used HuggingFace’s Transformer library combined with FastAI v2 (following the FastAI’s example\footnote{https://docs.fast.ai/tutorial.transformers}), as included on the code repository that supports this paper. To train the neural networks, we have used some techniques worth to mention (all the details are in the code repository). To find the most appropriate learning rate to use automatically, we used the function \textit{lr\_find} provided by FastAI following the proposal of \cite{smith2017cyclical}. This function trains the DNN over the dataset for a few iterations while varying from very low to very high learning rates at the beginning of each mini-batch of data to find which is optimal one regarding error (loss) metrics until the DNN diverges. To pursue a faster convergence, we schedule the learning rate as described in \cite{smith2019super} using the one cycle policy (\textit{fit\_one\_cycle}) in FastAI. Considering the transfer learning technique used, we trained the first “one cycle” on the top of the existing pre-trained model to later unfreeze all the model layers and do a more extended training (10-30 epochs) to improve the results.
Regarding other training details, in general, we used the default parameters from FastAI, except for a fixed multiplier to control all the dropouts (\textit{drop\_mult}) in AWD-LSTMs and AWD-QRNNs set to 0.3 because of some heuristics discovered during testing this research. Also, we decided to train similar architectures using a fixed number of epochs to make the models comparable. For the AWD-LSTM and AWD-QRNN, we used 30 epochs for fine-tuning because we found during the experimentation that the most remarkable improvement for every model produced occurs during that range of iterations. Similarly, we fine-tuned the transformers for ten epochs since we did not find a significant improvement after that. For more information about the training setup and software details, please refer to the repository that supports this paper and the FastAI documentation.

Finally, the hardware used to run the different software and neural networks training was a computer running Ubuntu Linux 18.04 LTS Bionic Beaver (64 bits). It has two Nvidia Tesla V100 GPUs x 16 gigabytes of RAM (Nvidia CUDA version 10.1), a CPU with 16 cores Intel(R) Xeon(R) CPU E5-2690 v4 @ 2.60GHz, 120 gigabytes of RAM, and 120 Gigabytes for the primary disk (HDD).

All the supporting materials and software details related to this paper are publicly available in a GitHub repository  \cite{software_repo_2020}. The NN models produced are under the Zenodo record \cite{models_software_repo_2020}.

\section{Results}
This section presents the results achieved after the full training of the selected DNN architectures with the different tokenization models.

As outlined in the previous section, we trained AWD-LSTM and AWD-QRNN DNN architectures using different tokenization models - word, unigram, BPE, and char-, and Transformer, using three different base models (GPT-2, BERT, and RoBERTa). We trained every AWD-LSTM and AWD-QRNN using one epoch to fit the model’s head and fine-tuned for 30 epochs. Meanwhile, the Transformer networks were trained equally for one epoch to fit the head and fine-tune the models for ten epochs.

We followed a two-way strategy to evaluate the NNs trained: use the NN training metrics and human evaluation of the models’ output. The metrics used are some of the most common in the literature: accuracy for the validation set and loss for the training and validation sets. They help the researchers understand how the NN acts over time, how the model is fitted to the dataset, and the performance and error scores while using training and validation datasets. In this case, the accuracy is the score concerning the LM’s ability to predict the next word of filling the missings accurately given a sequence of words from the validation set. The loss metrics report the error after applying the DNN to the training or validation set, respectively. Every implementation detail related to the DNNs and the metrics is available in the GitHub repository \cite{software_repo_2020}. Apart from those metrics, we assessed the models’ quality by applying them in the proposed tasks -generate text and auto-complete- and observing how they perform.

\subsection{Training results}

\begin{figure}[h]
\includegraphics[width=0.8\columnwidth]{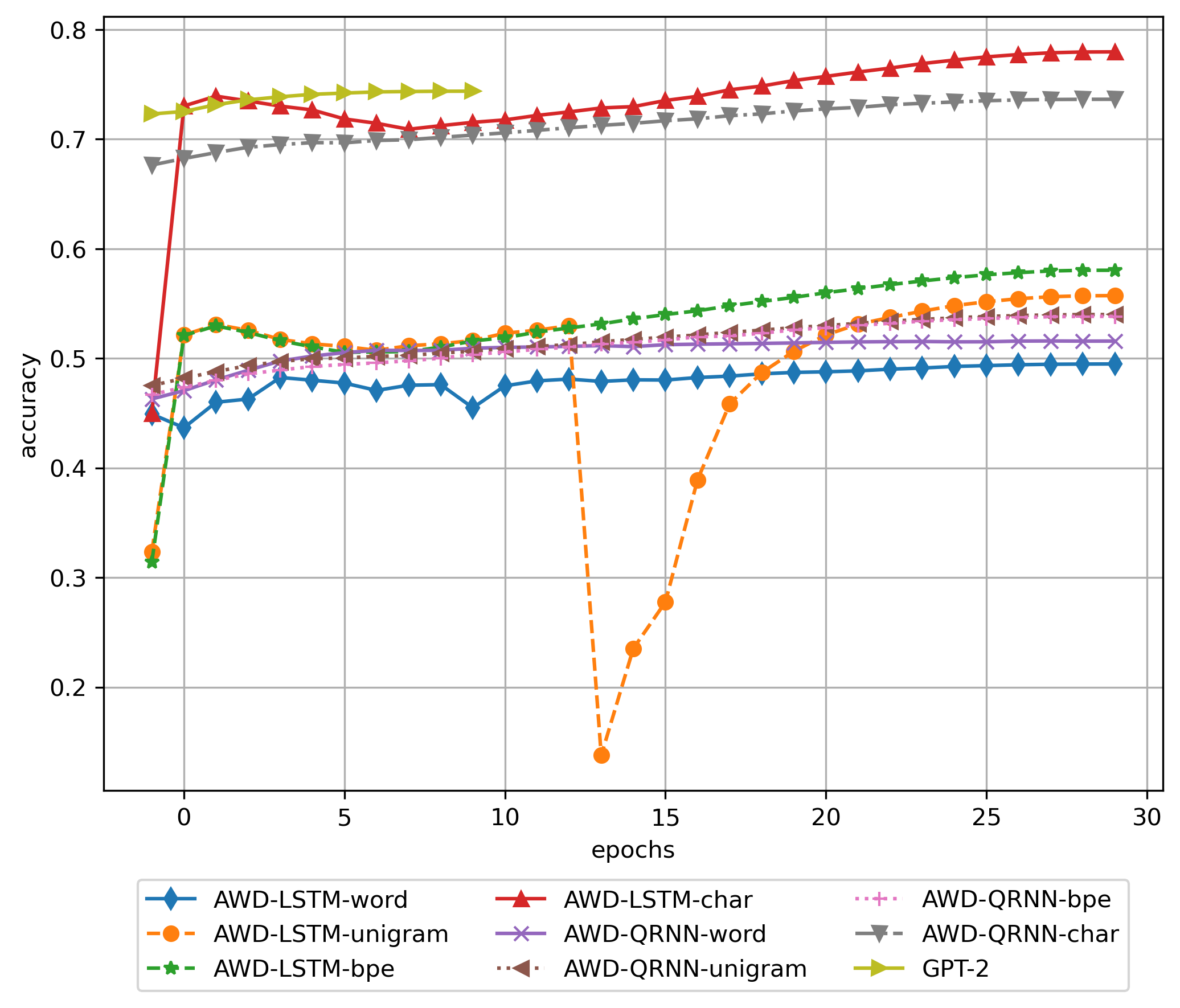}
\centering
\caption{Evolution of the accuracy of neural networks devoted to source code generation during the training epochs}
\label{evolution-accuracy-generate}
\end{figure}

\begin{figure}[h]
\includegraphics[width=0.8\columnwidth]{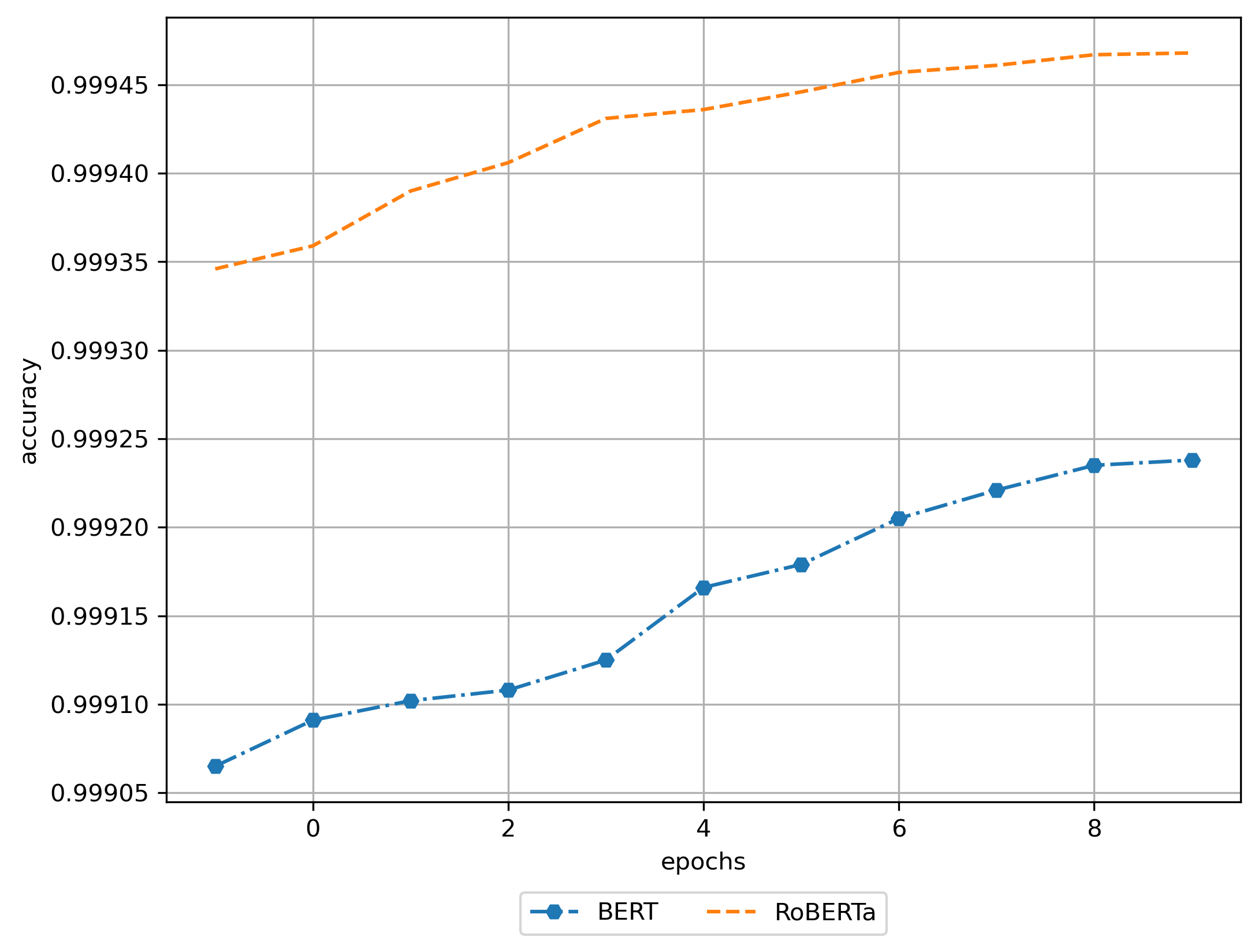}
\centering
\caption{Evolution of the accuracy of neural networks devoted to filling the blanks during the training epochs}
\label{evolution-accuracy-fill}
\end{figure}

\begin{figure}[h]
\includegraphics[width=0.8\columnwidth]{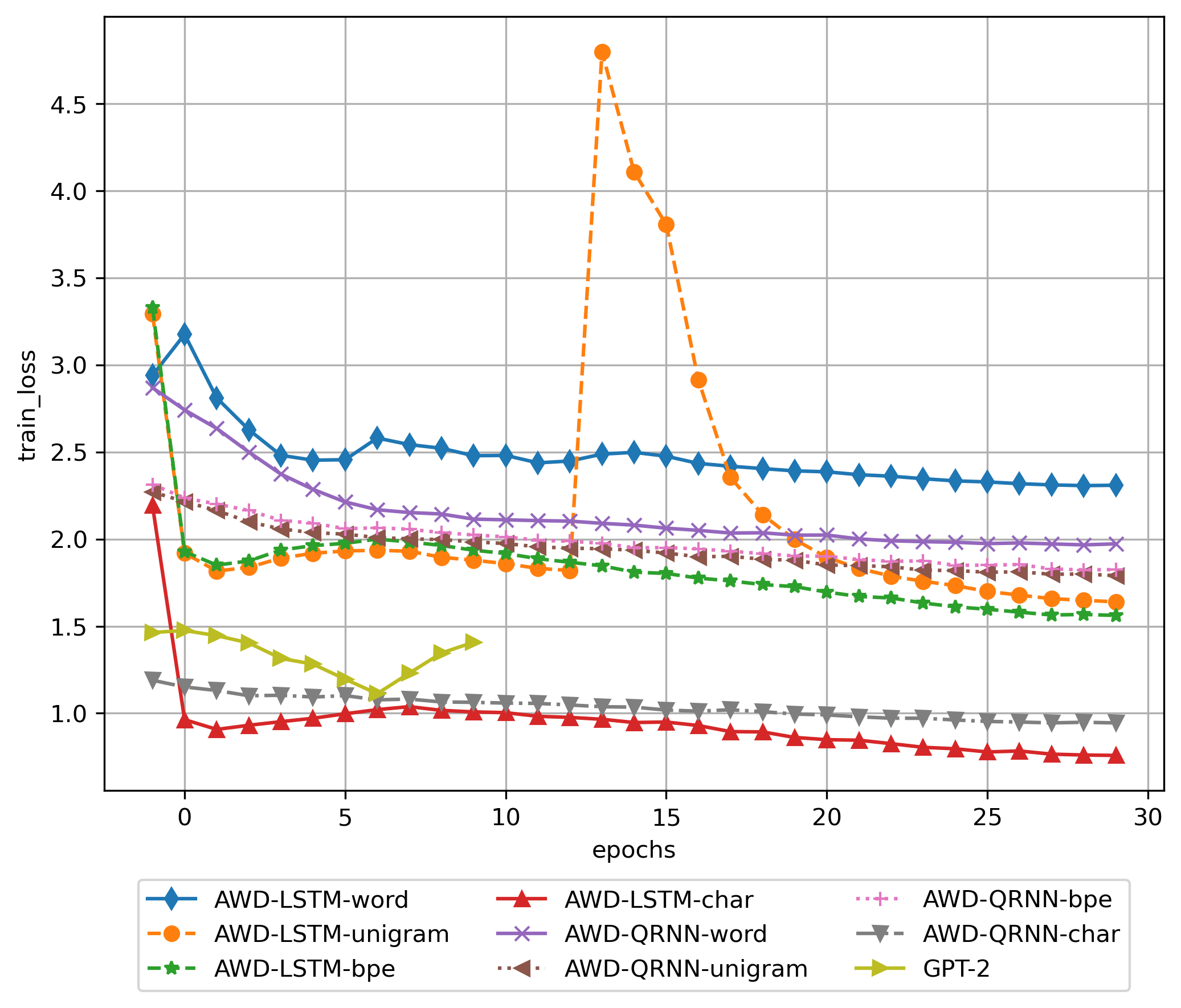}
\centering
\caption{Evolution of the \textit{training\_loss} of DNNs devoted to generating source code during the training epochs}
\label{evolution-training_loss-generate}
\end{figure}

\begin{figure}[h]
\includegraphics[width=0.8\columnwidth]{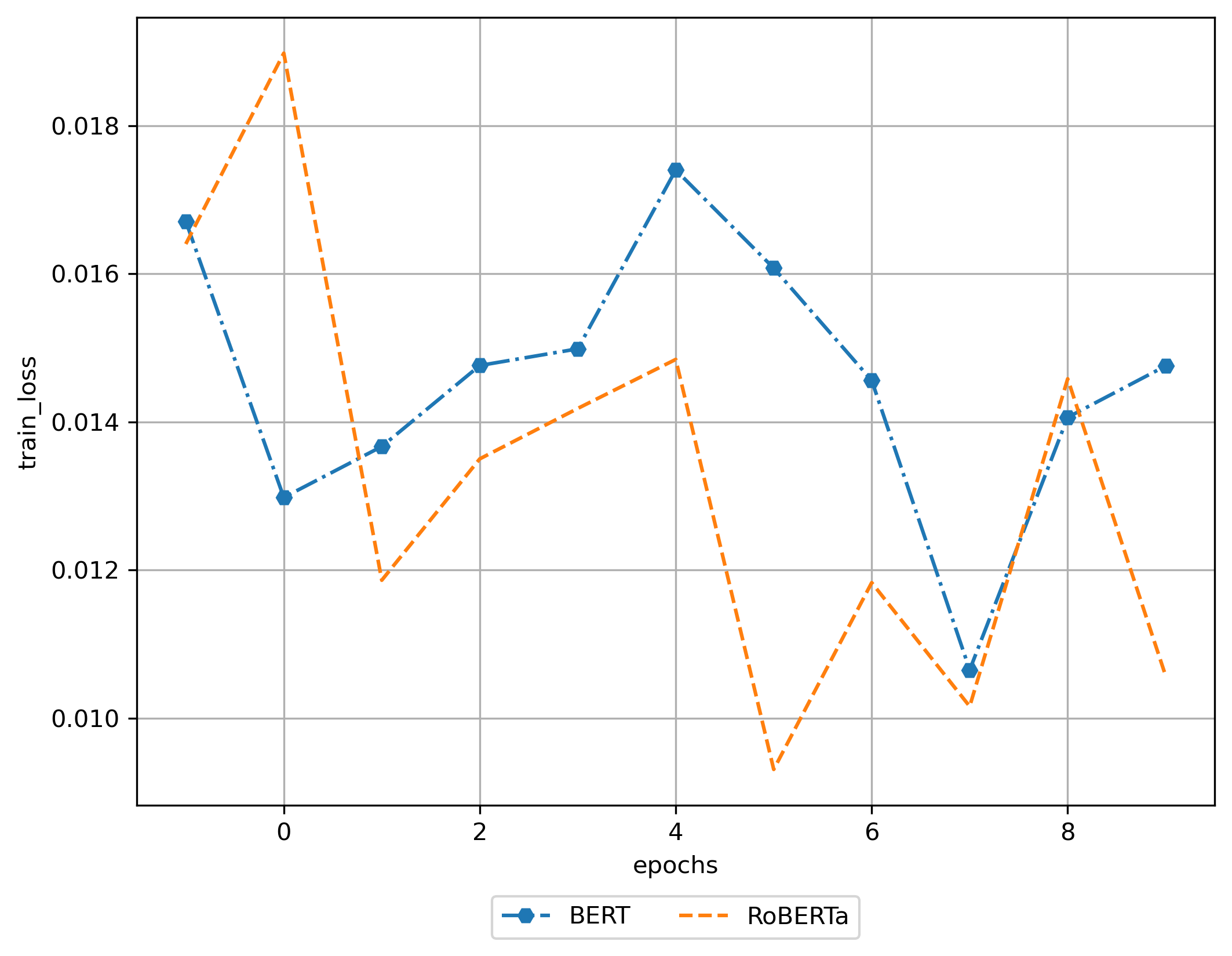}
\centering
\caption{Evolution of the \textit{training\_loss} of neural networks devoted to filling the blanks during the training epochs}
\label{evolution-training_loss-fill}
\end{figure}

\begin{figure}[h]
\includegraphics[width=0.8\columnwidth]{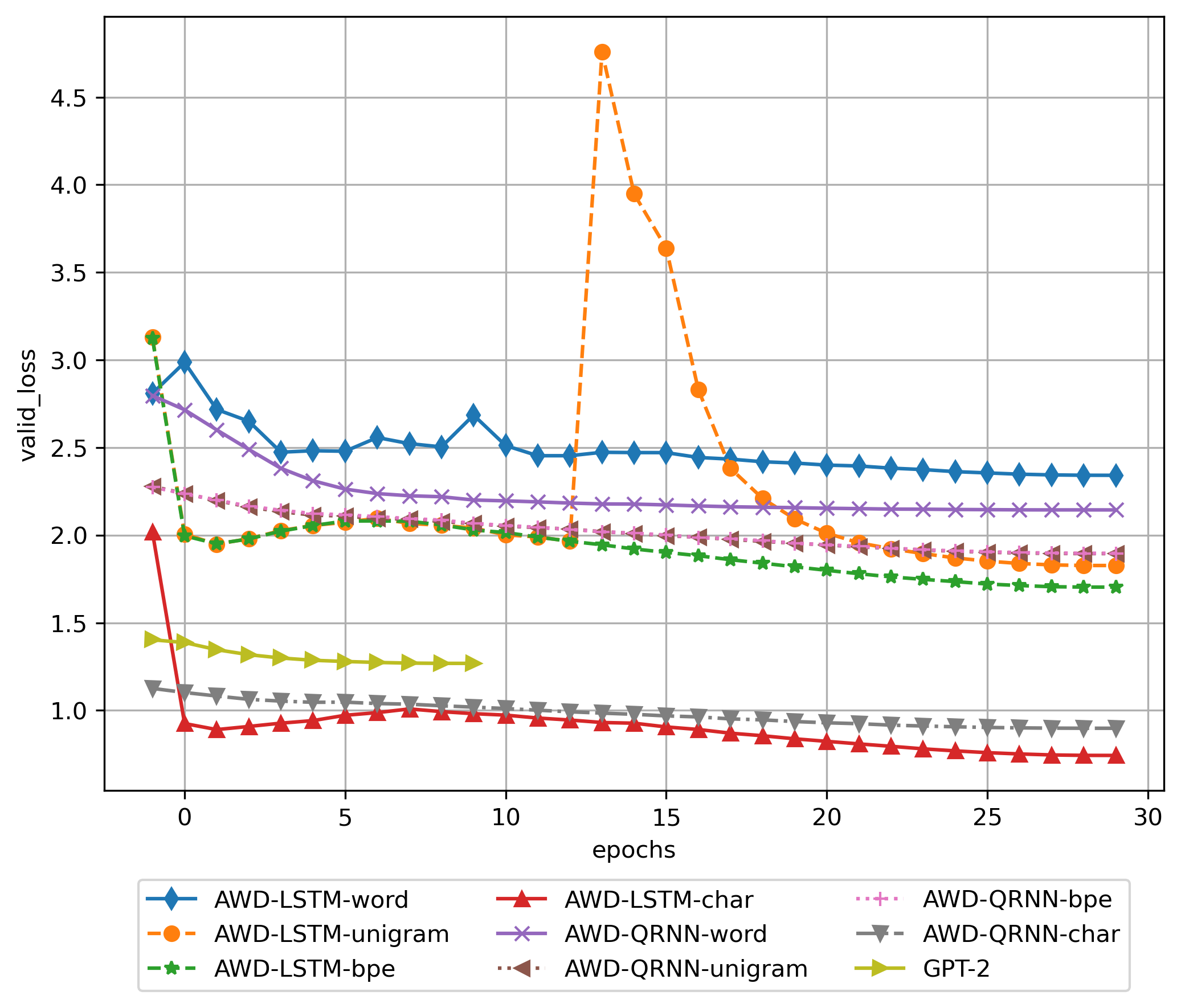}
  \centering
  \caption{Evolution of the \textit{validation\_loss} of DNNs devoted to generating source code during the training epochs}
  \label{evolution-validation_loss-generate}
\end{figure}

\begin{figure}[h]
\includegraphics[width=0.8\columnwidth]{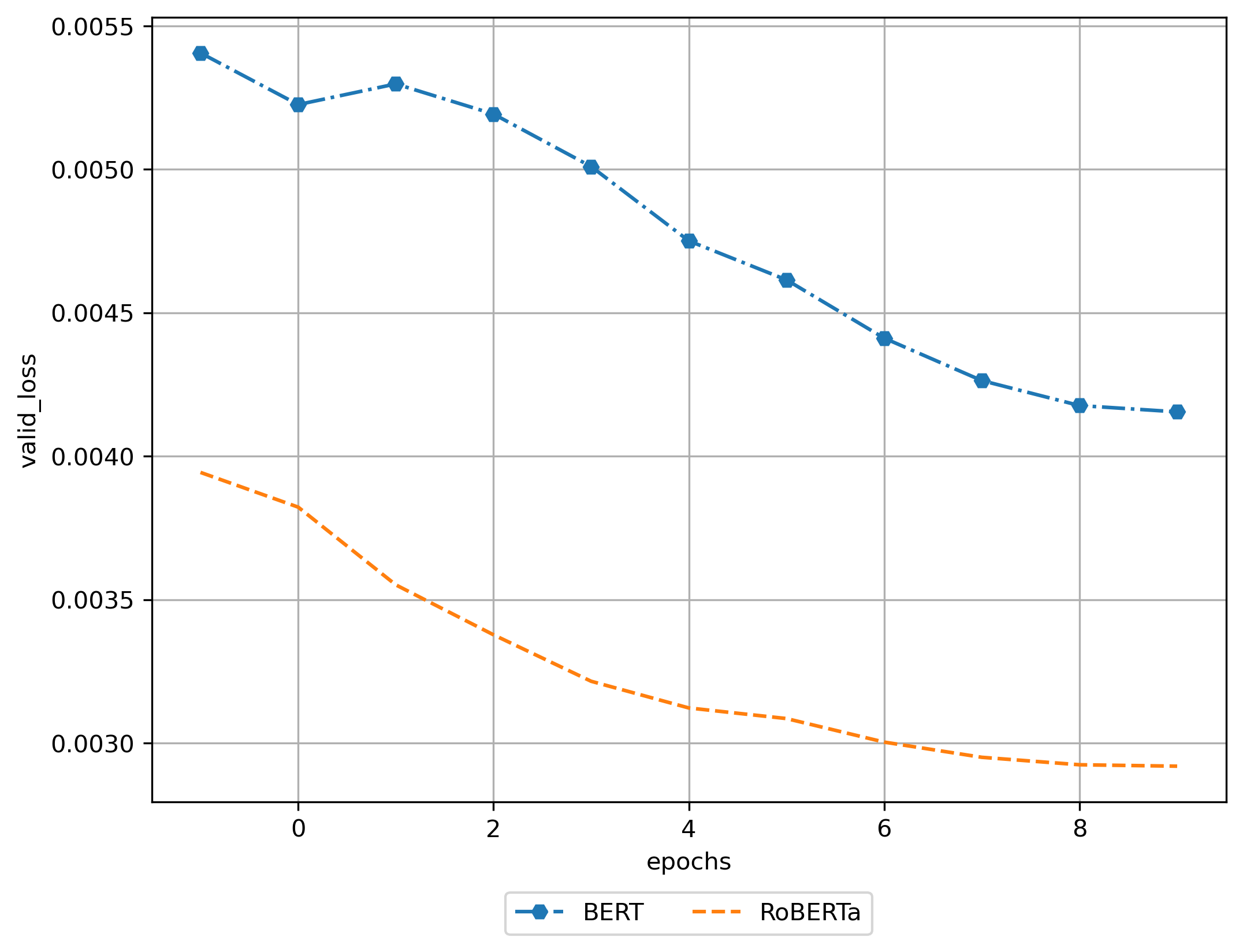}
\centering
\caption{Evolution of the \textit{validation\_loss} of neural networks devoted to filling the blanks during the training epochs}
\label{evolution-validation_loss-fill}
\end{figure}

Table \ref{final-results} displays the final metrics for the different NNs at the end of the training. Similarly, figures \ref{evolution-accuracy-generate} and \ref{evolution-accuracy-fill} show the evolution of each model’s accuracy during the training. Figures \ref{evolution-training_loss-generate}, \ref{evolution-training_loss-fill}, \ref{evolution-validation_loss-generate}, and \ref{evolution-validation_loss-fill} show the \textit{training\_loss} and \textit{validation\_loss} evolution along the training epochs.

\begin{table}[h]
\resizebox{0.95\columnwidth}{!}{
\begin{tabular}{l|c|c|c|c|c}
\centering
    \textbf{DNN architecture} & \textbf{Epochs} & \textbf{Accuracy} & \textbf{Train loss} & \textbf{Validation loss} & \textbf{Pre-trained}?\\
    \hline
    AWD-LSTM \textit{word}   &   31  &   0.494893  & 2.308937    &   2.341698    & Yes\\
    AWD-LSTM \textit{unigram}   &    31  &   0.557226  & 1.639875    &   1.826841    &   Yes\\
    AWD-LSTM \textit{BPE}   &    31  &   0.580373  & 1.561393    &   1.703536    &   Yes\\
    AWD-LSTM \textit{char}   &    31  &   0.779633  &    0.757956    &   0.742808    &   Yes\\
    \hline
    AWD-QRNN \textit{word}   &    31  &   0.515747  &    1.972508    &   2.144126    &   No\\
    AWD-QRNN \textit{unigram}   &    31  &   0.539951  & 1.790150    &   1.894901    & No\\
    AWD-QRNN \textit{BPE}   &    31  &   0.538290  & 1.824709    &   1.896698    & No\\
    AWD-QRNN \textit{char}   &    31  &   0.736358  &    0.944526 &  0.897850    &    No\\
    \hline
    GPT-2   &   11  &   0.743738  & 1.407818    &   1.268246    &   Yes\\
    BERT    &   11  &   0.999238  & 0.014755    &   0.004155    &   Yes\\
    RoBERTa &   11  &   0.999468    &   0.010569    &   0.002920    &   Yes\\

\end{tabular}
}
\caption{Results after full training of each NN architecture}
\label{final-results}
\end{table}

On the one hand, according to the results displayed in Table \ref{final-results} and Figure \ref{evolution-accuracy-generate}, for neural networks intended for automated source code generation -AWD-LSTM, AWD-QRNN, and Transformer GPT-2-, the overall NN-tokenization model combination performed better in the case of accuracy metrics was the AWD-LSTM with char tokenization (accuracy 0.779633). The second one was the GPT-2 transformer model -BPE over raw bytes tokenization- (0.743738), and the third one the AWD-QRNN with char tokenization (0.736358). Related to AWD-LSTM and AWD-QRNN architectures’ combination with other tokenization techniques, we obtained poor results on accuracy: between 0.494893 to 0.580373. On the other hand, according to the results shown in Table \ref{final-results} and Figure \ref{evolution-accuracy-fill}, both models (BERT and RoBERTa) had excellent accuracy results in the transformer models intended for auto-completion 0.999238 and 0.999468, respectively.

About how the pre-training and transfer learning affects the results, the two top results regarding the accuracy come up from pre-trained models in the English language (0.779633 and 0.743738), yet the third-best result was from a non-pre-trained network (0.736358). Comparing the similar networks, the average (mean) accuracy of the AWD-LSTM pre-trained versions is 0.603031 (standard deviation -std- of 0.123144), while the average accuracy of AWD-QRNN non-pre-trained versions was 0.582587 (std of 0.103107). The only combination NN-tokenization model that worked worse when it was pre-trained was the one with the word tokenization.

Regarding the observed losses, it is worth commenting that the AWD-LSTM \textit{char}, AWD-QRNN \textit{char}, and the three transformer models (GPT-2, BERT, RoBERTa) could be trained for more epochs or with a higher learning rate. The model may be be underfitting since the training loss is higher than the validation loss (Table \ref{final-results}, figures \ref{evolution-training_loss-generate}, \ref{evolution-training_loss-fill}, \ref{evolution-validation_loss-generate}, and \ref{evolution-validation_loss-fill}).

To put in context the accuracy achieved during the experimentation, we compare the results with the existing literature. The papers \cite{tiwang2019deep,li2017code,raychev2016probabilistic} present models trained to generate Python code considered state of the art when published. Our three best models trained to generate source code outperform the approaches based on the following architectures: Vanilla LSTM (accuracy: 67.3\%) \cite{li2017code}, Attention LSTM (69.8\%) \cite{li2017code}, Pointer Mixture Network (70.1\%) \cite{li2017code}, or Probabilistic Model (69.2\%) \cite{raychev2016probabilistic}. Our models perform worse than other approaches based on the abstract syntax tree (AST) instead of using the textual code: AST-LSTM AST-LSTM (accuracy 90.3\%) \cite{tiwang2019deep} or AST-MLP (90.1\%) \cite{tiwang2019deep}. So, considering this, our results are good in terms of accuracy compared to the existing peer-reviewed literature. As a side note, we did not find models available publicly to compare with, neither in these nor in other peer-reviewed papers. For the models devoted to auto-complete code (BERT and RoBERTa), as of now, we did not find peer-reviewed papers trying to solve this task using Transformer architectures.

\subsection{Results of using the trained DNN models}

Following, we introduce some examples of how the different neural networks generated source code or auto-completed it. These examples are used to assess the model’s output quality, avoiding relying only on the NN metrics. This validation using more methods than the simple metrics is common in the AI and NLP fields. Many papers employ human evaluation or specialized metrics to grasp better the model’s performance and capture any other particular detail or quality indicator on the related task proposed. In this case, we show only those with the best accuracy results. To generate source code, we show output from AWD-LSTM \textit{char}, AWD-QRNN \textit{char}, and GPT-2. For auto-completion, we show results from BERT and RoBERTa. For the source code generation, every model faced the same challenges \cite{software_repo_2020}. Those are the results when any of them try to generate five different sentences with the input \textit{from pandas import}:

\begin{quote}{AWD-LSTM \textit{char}}
\begin{scriptsize}\begin{verbatim}
"from pandas import defaults as _rp pandas ="
"from pandas import DataFrameEncoder if self"
"from pandas import OrderedDict else: passed"
"from pandas import sequence if type(seq1) i"
"from pandas import conditional_insertion as"
\end{verbatim}\end{scriptsize}
\end{quote}

\begin{quote}{AWD-QRNN \textit{char}}
\begin{scriptsize}\begin{verbatim}
"from pandas import webbrowser self.login fr"
"from pandas import __http_args or [ 'Pipeli"
"from pandas import iterations is not None p"
"from pandas import service params to servic"
"from pandas import difference if not isinst"
\end{verbatim}\end{scriptsize}
\end{quote}

\begin{quote}{GPT-2}
\begin{tiny}\begin{verbatim}
"from pandas import time, np\n
                      "
"from pandas import pandas from time.time.datetime import Date\n
             with n"
"from pandas import gtk, os\n
    from pandas_utils import pandas_utils\n
    import pylint"
"from pandas import wcpy\n
        import cpy_context as cpy_context\n
    "
"from pandas import gkpy\n
     """\n
     ... pass\n
    kwargs = cg"
\end{verbatim}\end{tiny}
\end{quote}

To assess the generation, we do not focus on the semantics of the imports used or whether they are part of the Pandas library or not, but on the language’s correctness. In general, from a Python language perspective, the outputs from GPT-2 are better. The outputs include line breaks, indentation, and fair use of multiple inputs in one sentence (except in one of the outputs). The AWD-LSTM and AWD-QRNN failed to auto-generate an import sentence appropriately, or at least, they fail regarding the usual manner used by regular users. As a final comment on this, the other models trained failed on similar issues, plus they do not get enough semantic context related to the Pandas library.

Similarly, concerning the source code auto-completion, both BERT and RoBERTa tried to autofill the \textit{mask} token in the sentence \textit{from pandas import [MASK]}. These are the results:

\begin{quote}{BERT}
\begin{tiny}\begin{verbatim}
[{'sequence': '[CLS] from pandas import [SEP] [SEP]',
  'score': 0.9969683289527893,
  'token': 102,
  'token_str': '[SEP]'},
 {'sequence': '[CLS] from pandas import [CLS] [SEP]',
  'score': 0.0010887219104915857,
  'token': 101,
  'token_str': '[CLS]'},
 {'sequence': '[CLS] from pandas import. [SEP]',
  'score': 0.0004200416151434183,
  'token': 119,
  'token_str': '.'},
 {'sequence': '[CLS] from pandas import ; [SEP]',
  'score': 0.00027348980074748397,
  'token': 132,
  'token_str': ';'},
 {'sequence': '[CLS] from pandas import def [SEP]',
  'score': 8.858884393703192e-05,
  'token': 19353,
  'token_str': 'def'}]
\end{verbatim}\end{tiny}
\end{quote}

\begin{quote}{RoBERTa}
\begin{tiny}\begin{verbatim}
[{'sequence': '<s>from pandas import\n</s>',
  'score': 0.6224209666252136,
  'token': 50118,
  'token_str': 'Ċ'},
 {'sequence': '<s>from pandas import.</s>',
  'score': 0.22222988307476044,
  'token': 4,
  'token_str': '.'},
 {'sequence': '<s>from pandas import </s>',
  'score': 0.038354743272066116,
  'token': 1437,
  'token_str': 'Ġ'},
 {'sequence': '<s>from pandas import\n\n</s>',
  'score': 0.028566861525177956,
  'token': 50140,
  'token_str': 'ĊĊ'},
 {'sequence': '<s>from pandas import.</s>',
  'score': 0.021909384056925774,
  'token': 479,
  'token_str': 'Ġ.'}]
\end{verbatim}\end{tiny}
\end{quote}

For auto-completion, we observe that almost no auto-completion is right. The more accurate sentences from the Python language point are the ones in which the mask token has been replaced by white space or by a dot. Nevertheless, they are not correct but closer to be right compared to the other ones. One interesting thing is that BERT assigns a very high score to a predicted mask, which is not correct, and shallow scores to the other possible solutions (also incorrect). In the case of RoBERTa, it gives lower scores to all the solutions, yet also fails on the correctness: the second sentence predicted (score 0.222) can be closer to be right compared to the first one (score 0.622).

\section{Discussion}

Considering the results obtained, one could convincingly assert that the tokenization model used profoundly affects the results when generating automated source code. Although that may be accurate, we must discuss it carefully.

\subsection{Discussing the outcomes from resulting models}

First, our overall results are consistent with the existing literature \cite{karampatsis2019maybe,ganin2016domain,kimcharacter2016,karpathy2016}. Sub-word tokenization works better in the case of modeling source code, as \citeyear{karampatsis2019maybe} stated. Every result obtained is consistent in that sense. Even more, as \citeyear{karpathy2016} envision, char tokenization probably should the best option to try by default when dealing with LMs and source code. Furthermore, according to the results achieved, models such as GPT-2 -using a tokenization model based on BPE over raw bytes- can outperform LSTM/QRNN models like those we tested to grasp better the internals of a programming language. As showcased during the results, even if GPT-2 was not the best model in terms of accuracy, it gave better code outputs than the other ones selected for the comparison.

As future work, it would be great to check if the better textual output in the case of GPT-2 is because of a) it is a much bigger and better pre-trained model (163,037,184 parameters against 36,491 for AWD-LSTM and AWD-QRNN models), b) it is related to dataset's size or quality, c) if it is related to both causes or, d) if it is related to other issues.

Continuing with the comments about the accuracy, one may note that the textual outputs generated by AWD-LSTM \textit{char}, AWD-QRNN \textit{char}, and GPT-2 could be polished to be more accurate. The final training loss is higher than the validation loss for the three selected DNN architectures, which can be a sign of underfitting. We find the same issue (\textit{train\_loss} << \textit{valid\_loss}) for BERT and RoBERTa-based models. Whether the purpose of this paper is not to produce state-of-the-art results \textit{per se}, we continued the training for over five more epochs to verify it. The improvement obtained from extending the training for the best approaches was residual in general, so we decided to report the results for 1+30 epochs in AWD-LSTMs and AWD-QRNNs, and 1+10 epochs for Transformer.

\subsection{Examining the effect of pre-training and transfer learning}

Regarding the pre-training and transfer learning, every pre-trained model (on English-based datasets) got better accuracy than its non-pre-trained counterparts except in word tokenization models. It seems to be strongly related to the statements we introduced at the beginning of this paper, citing \cite{allamanis2015suggesting, karampatsis2019maybe} about the source code does not have the same restrictions in the sense of common words or neologisms. In this sense, the conclusion comes into our mind rapidly: if we consider source code words in “word” units, they probably will not fit in the fixed set of words used in a human language like English. So, the LM’s knowledge acquired during the pre-training is not entirely valid when we get out of that fixed set of words that compose a language. Most words in the programming language are neologisms for the LMs pre-trained in English, and thus, it needs to incorporate them and their relationships into the learned knowledge. For the sub-word units, the LM can be less sensible to the neologisms.
Potentially, it could be more robust the more divided a word is since the set of bytes or chars is more straightforward than the chunks present in richer constructions or information units.

Going deeper into this research, concerning the pre-training effect over LMs modeling source code, it could be worth researching the relationship between the pre-training in different human-spoken languages and the LM ability to work with existing source code specific programming languages.

\subsection{Reviewing the textual assessment of resulting LMs}

About the tests made generating source code or filling in the blanks using the trained LMs, we think that, in general, the textual results obtained are not so good, yet they are informative of how LMs are working and how they can be improved. One of the things that can explain these results is the dataset used. In this case, we used a public dataset that other researchers can use to make results and experiments comparable and replicable. In the literature, we do not find a standard dataset for these tasks against which we can compare easily. Other papers \cite{tiwang2019deep,li2017code,raychev2016probabilistic} use custom datasets, but we find a lack in the literature of well-recognized code datasets to use. Comparing with other recent papers in the NLP field used as the basis for this research \cite{radford2019language, devlin2019bert, liu2019roberta, brown2020language}, the dataset may be relatively small to train a big LM to accomplish appropriately challenging tasks like generating source code or auto-completing it. Future work may be testing these or new approaches in bigger datasets to train big LMs focused on modeling the Python language and checking whether the results are better. Recent examples of LMs -such as GPT-3 \cite{brown2020language}- claim to produce accurate textual outputs even in contexts in which they were not trained. Part of the explanation given for that ability is the use of gargantuan datasets combined with Transformer and other attention-based architectures. So, those approaches can also be relevant to other contexts like ours. 
Another line for future research can be using datasets focused on specific libraries or Python aspects and verify if these approaches specialize positively for those contexts the DNN models used in this paper.

Related to evaluating the code generated or filled, we observed in the literature different approaches \cite{celikyilmaz2020evaluation, roziere2020unsupervised}. In the context of LMs modeling source code, many papers and software libraries devoted to translating between programming languages typically evaluate text generation using methods and metrics like BLEU \cite{papineni2002bleu}, or variants like SacreBLEU \cite{post-2018-call}. Other papers like \cite{tiwang2019deep} rely on the accuracy to assess an LM’s performance based on deep learning. Some models can even solve different tasks that are part of existing benchmarks \cite{weston2015towards} or are evaluated, checking their perplexity (similarly to those that evaluate the model using the accuracy).
The current tendency in large models is to evaluate them using human intervention to evaluate the output’s quality \cite{radford2019language,brown2020language}. We assessed the models using accuracy during our experiments and evaluated the models’ textual outputs based on our prior human knowledge. It would be interesting for the future to plan new evaluation processes involving larger cohorts of source code experts to evaluate the models such as \cite{radford2019language,brown2020language} do. One of the potential new assessments can be usability tests conducted with programmers. They can compare the code they would write against the code proposed by any of the DNNs presented here and the result from other common code auto-completion tools included in integrated development environments.
As we outlined in the results section, relying only on metrics like accuracy should not be enough. As in our case, accuracy and the other metrics can be a good indicator of the model’s performance, yet we need to verify LMs behavior and quality using complementary methods like specialized metrics or human evaluation. For tasks like auto-completion or source code generation, there are no existing specialized metrics (like BLEU in translation, for example), so one of the future research lines is improving the evaluation of LMs for source code. Based on some existing ideas in broad NLP, there are many opportunities to explore in that sense. From new test suites for language models used in source code contexts \cite{roziere2020unsupervised} to behavioral testing \cite{ribeiro-etal-2020-beyond} or human-centric evaluation of the models \cite{celikyilmaz2020evaluation} with particular emphasis on reproducible and unbiased assessments, or combinations of automatic testing and human-centric assessments.

\section{Conclusions}

This paper compares how different approaches to tokenization models, deep neural network architectures, pre-trained models, and transfer learning affect the results from language models used to generate source code or auto-complete software pieces. We studied different DNN architectures like AWD-LSTM, AWD-QRNN, and Transformer to seek which kind of them work better with different tokenization models (word, unigram, BPE, and char). Also, we compared the pre-training effect on the results given by LMs after training them and fine-tuning them via transfer learning to work with other languages (English language to Python programming language). As a result of this work, we find that in small LMs (like our AWD-LSTM and AWD-QRNN models), the tokenization using char-sized chunks works better than using any other tokenization models. In larger models like the Transformer GPT-2, the accuracy was slightly worse than the other architectures. However, GPT-2 raised better results on the source code generation tests (even using another tokenization approach like BPE over raw bytes). For source code auto-completion, we tested some transformer models like BERT and RoBERTA. While their accuracy was above any other models, they did not perform very well when performing the tasks proposed in our tests. In general, we find that pre-trained models work better, even if they were not trained initially for a programming language like Python (our models were pre-trained using the English language). Finally, related to evaluating tasks like automating source code generation and source code auto-completion, we raise concerns about the literature gaps and propose some research lines to work on in the future.

\section{ Acknowledgments}
We thank the IBM Quantum team and the IBM Research ETX team for the insightful discussions about this research and the support received during the development of this research.

\bibliography{bibliography.bib}

\end{document}